%% file: cvpr.tex
\documentclass[final]{cvpr}

\usepackage{times}
\usepackage{epsfig}
\usepackage{graphicx}
\usepackage{amsmath}
\usepackage{amssymb}

\usepackage[table]{xcolor}
\usepackage{multirow}
\usepackage[pagebackref=true,breaklinks=true,colorlinks,bookmarks=false]{hyperref}

\def\cvprPaperID{2} 
\def\confYear{CVPR 2021}

\begin{document}

\title{Zero-shot Learning with Class Description Regularization}

\author{Shayan Kousha\textsuperscript{1}, Marcus A. Brubaker\textsuperscript{1, 2}\\
\textsuperscript{1}Department of Electrical Engineering and Computer Science, York University, Toronto, Canada\\
\textsuperscript{2}Vector Institute, Toronto, Canada\\
{\tt\small [shayanko, mab]@eecs.yorku.ca}
}

\maketitle

\begin{abstract}
\end{abstract}
 The purpose of generative Zero-shot learning (ZSL) is to learning from seen classes, transfer the learned knowledge, and create samples of unseen classes from the description of these unseen categories. To achieve better ZSL accuracies, models need to better understand the descriptions of unseen classes. We introduce a novel form of regularization that encourages generative ZSL models to pay more attention to the description of each category. Our empirical results demonstrate improvements over the performance of multiple state-of-the-art models on the task of generalized zero-shot recognition and classification when trained on textual description-based datasets like CUB and NABirds and attribute-based datasets like AWA2, aPY and SUN.
   Code is available at \textit{https://github.com/shayan-kousha/DGRZSL}
   
\section{Introduction}
Image classification methods have advanced significantly in the past few years.
This has largely been driven by a large amount of data per class which has enabled models to learn them.
However, data gathering can be time-consuming and expensive.
Further, many rare classes may not have sufficient training data. This has led to the creation of ``Zero-Shot Learning'' (ZSL) methods which aim to leverage other information, typically natural language descriptions of classes, to learn about classes with little or no directly labelled data available. Recent generative ZSL methods have gone further; instead of only classifying unseen classes, they aim to also generate samples from unseen classes \cite{ijcai2017-246, Yizhe_ZSL_2018, xian2018feature, long2017zeroshot, 7907197, Mohamed_Creative_2019}.

Despite recent advances in the field of generative ZSL, there are still significant challenges.
Generative ZSL methods do not guarantee that the generated visual examples of unseen classes deviate meaningfully from seen classes.
That is, there is a risk that the generated images are too similar to samples from the seen classes.
Another problem arises when the model is forced to generate samples of unseen classes that arbitrarily deviate from seen classes.
In this case there is a risk that the generated images do not follow the description of unseen classes.
Instead, the primarily property of the generated images is that they deviate sufficiently from seen classes.

We believe paying closer attention to the details of the description is key to solving both of these issues.
Inspired by this, we introduce a new model that aims to encourage the generative model to pay closer attention to the details.
Specifically, the model includes a mapping from the generated visual features back to the original text or attributes of a class.
By requiring that there exists a mapping from the generated visual features to the class specific description, we force the generator to pay closer attention to these inputs.
This is implemented by adding an additional loss function which penalizes the generator and regularizer if the generated description is not similar to the input description.
We call our proposed method Description Generator Regularized ZSL (DGRZSL). 
The approach is unsupervised and not tied to a specific generative ZSL approach so it can be added to any ZSL approach that uses the descriptions of seen and unseen categories with minimal modifications to the underlying generative ZSL approach.

\input{figs/fig_model}

\section{Setting} 
\label{setting_sec}
In our zero-shot learning setting each data point consists of visual features, a class label, and a semantic representation of the class. These semantic representations are either textual or attribute based. In this section, we introduce notations to represent training and test data.

Let $r_{i}^{s} \in \tau$ and $r_{i}^{u} \in \tau$ represent semantic representations of seen and unseen classes where $\tau$ is the semantic space from distribution $p_{rep}$. $N_{s}$ is the number of seen (training) image examples, $x_{i}^{s} \in X$ is the visual features of the $i^{th}$ image in the visual space $X$ from distribution $p_{data}$, and $y_{i}^{s}$ is the corresponding category label. The available training data is denoted as $D^{s} = {(x_{i}^{s}, y_{i}^{s}, r_{i}^{s})}_{i=1}^{N^{s}}$ where we have $K^{s}$ unique seen class labels. Additionally, we denote the set of seen and unseen class labels as $S$ and $U$ where $S$ and $U$ do not have any labels in common. Then the zero-shot learning task is formulated as predicting the label $y^{u} \in U$ of an unseen class sample $x^{u} \in X$. Generalized ZSL (GZSL) is formulated, as  predicting the label of $y \in U \cup S$ which means the search space at test time includes labels from both seen and unseen classes \cite{xian2020zeroshot}.

\section{Method}
\label{method_sec}
Figure \ref{fig:model} shows an overview of our model which we describe next.
The basic generative ZSL model is based on a generative adversarial network  \cite{goodfellow2014generative} and was introduced in \cite{Yizhe_ZSL_2018}.
A generator network is trained to map samples of noise and a representation of the class into visual features.
The noise, $z$, is sampled from mean zero standard deviation 1 Gaussian distribution.
A discriminator network takes as input visual features.
Its output is a classification as to whether the input visual features were real or were generated.
The discriminator network can also include an additional output head which predicts the class label.
The real/fake prediction of the discriminator for an input image and the predicted label of a seen class $k \in S$ given the image are defined as $D^{r}(.)$ and $D^{s,k}(.)$, respectively.
The architectures for these networks are as described in \cite{Yizhe_ZSL_2018}.

The contribution of this work is the addition of a semantic representation generator network (SR) and corresponding loss to this model.
The SR generator network learns to map from the visual features of a sample to the semantic representation of the class.
An added loss function (described below) penalizes differences between the output of the SR network and the provided semantic representation of the class.
This explicitly requires the generated visual features to contain more information about the semantic representation of a class and encourages better generalization to the unseen classes.
The SR generator network consists of three fully connected layers accompanied with ReLU activations to generate semantic representations that describe the input visual features.
We explain the loss function in detail in the following sections.

Previous work \cite{Mohamed_Creative_2019} identified a challenge with training generative ZSL models of this form.
Specifically, the generator, $G$, never sees data from the unseen classes, neither visual features nor semantic representations.
While this is, of course, the definition of zero-shot learning, it means that the generator sees very limited variability in semantic representations during training.
In response \cite{Mohamed_Creative_2019} proposed augmenting the training process to include novel, hallucinated semantic representations of new classes which the generator would try generate samples for.
To generate the new semantic representations, two classes $a$ and $b$ are picked at random and with $r_{a}^{s}, r_{b}^{s} \in \tau$ denoting their semantic representation.
Then a random convex sum of these features to used create the hallucinated representation:
\begin{equation}
    r^{h} = \alpha r_{a}^{s} + (1 - \alpha) r_{b}^{s}
\end{equation}
where alpha is uniformly sampled from interval [0.2, 0.8]  \cite{Mohamed_Creative_2019}.

After training, the generative model can be used to generate visual features of unseen classes.
These samples can then be used to train a classifier as in a regular, classification task

\subsection{SR Loss Function}
Here we introduce the DGRZSL loss.
This loss is in addition to other terms that are commonly used in existing generative ZSL approaches \cite{xian2018feature, Yizhe_ZSL_2018, Mohamed_Creative_2019}.
The main contribution of DGRZSL is the addition of a Semantic Representation (SR) network.
The SR network maps from visual features to semantic representations.
To constrain the output of this network, and encourage the visual features to represent information present in the semantic features, the added loss function encourages the generated semantic representation for the generated features to match the input.
In essence, the model ensures that the combination of the visual feature generator and the semantic representation generator form an autoencoder.
The loss for our SR generator network is as follows:
\begin{equation}
\begin{aligned}[b]
L_{SR} =
& - \mathbb{E}_{x,r^{s} \sim p_{data}}[sim(r^{s}, SR(x))] \\
& - \mathbb{E}_{z \sim p_{z}, r^{s} \sim p_{rep}^{s}}[sim(r^{s}, SR(G(r^{s}, z)))]
    \\
& - \mathbb{E}_{z \sim p_{z}, r^{h} \sim p_{rep}^{h}}[sim(r^{h}, SR(G(r^{h}, z)))]
\end{aligned}
\end{equation}
where $x$ denotes the visual features, $r^{s}$ denotes the semantic representations of the seen classes, $p_{data}$ is the training data distribution of $x,r^{s}$, $r^{h}$ denote the hallucinated semantic representations generated from $p_{rep}^h$ as described above, and $sim$ is a function which measures the similarity between semantic representations.
All terms encourage the semantic representations produced by $SR(\cdot)$ to be similar to the ``correct'' semantic representation, even in the case of hallucinated semantic representations.
The first term considers visual features from the training data with known classes and hence known semantic representations.
The second term considers generated visual features given semantic representations of known classes.
Finally, the third term considers generated visual features with hallucinated semantic representations.
While all terms encourage $SR$ to generate accurate semantic representations from the visual features, most critically, the last two terms also encourage $G$ to produce visual features which meaningfully capture the input semantic representations.

\subsection{Training}
The above SR loss function is used in conjunction with the standard generator and discriminator losses and adversarial training used in GAZSL \cite{Yizhe_ZSL_2018} and CIZSL \cite{Mohamed_Creative_2019}.


\input{figs/tab_attribute_results_both_evaluation_methods}
\input{figs/tab_textual_results_both_evaluation_methods}

\section{Experiments}
\label{experiments_sec}
We evaluate our method on textual-based datasets including \textit{Caltech UCSD Birds-2011} (CUB) \cite{WahCUB_200_2011} and \textit{North America Birds} (NAB) \cite{7298658}, which are split into easy and hard benchmarks by \cite{elhoseiny2017link}, and attribute-based datasets including AWA2 \cite{xian2020zeroshot}, SUN \cite{conf/cvpr/PattersonH12}, and APY \cite{5206772}.
The metrics we considered in our experiments are Top-1 unseen class accuracy, Seen-Unseen Generalized Zero-shot performance with area under Seen-Unseen curve \cite{chao2017empirical} (Seen-Unseen AUC), and Harmonic mean \cite{xian2020zeroshot} (Seen-Unseen H).

\input{figs/fig_AUC}

\subsection{Baselines and Evaluation}
The most relevant baselines for our methods are the
GAZSL \cite{Yizhe_ZSL_2018} and CIZSL \cite{Mohamed_Creative_2019} ZSL models on which our approach is built.
These methods are state-of-the-art generative ZSL approaches.
Both GAZSL \cite{Yizhe_ZSL_2018} and CIZSL follow the same procedure to evaluate the performance of their models.
A dataset is split into training and test sets.
The training set is used to train the weights of the networks and performance on the test set is evaluated every few epochs. 
The final reported performance is of the model which achieved the best test set accuracy during training.
A similar procedure was used to tune hyperparameters.
However, this is an unrealistic representation of model performance as model selection is done based on the test set itself.
Instead, we propose to use a validation set, disjoint from the training and testing sets, to select the final model.
After training, the performance on this validation set is used to select a model and evaluate the performance on the test set to give a more fair and accurate picture of model performance.
However, as a consequence, the results reported for the baseline models when our evaluation method is used differ from the results reported in their original papers \cite{Mohamed_Creative_2019, Yizhe_ZSL_2018}.
For transparency, we report results with both evaluation protocols.
Our model outperforms both baseline models in most cases regardless of evaluation methods used.
In what follows we limit analysis to the results obtained by using the validation set for model selection.

\subsection{ZSL Recognition Results}
\label{results_sec}
Table \ref{tab:attribute_results_combined} summarizes the accuracy achieved of the proposed method and the two baseline models on the attribute-based datasets.
DGRZSL outperforms the state-of-the-art baseline methods in top-1 accuracy in all cases with an average improvement of 9.4\% in the case of the APY dataset.
DGRZSL also displayed significantly improved performance in the seen-unseen H metric, improving it by 29.65\%, 1.63\%, and 10.91\% over state-of-the-art on AWA2, SUN, APY, respectively. 

Table \ref{tab:textual_results_combined} shows the results achieved by our model on CUB and NAB datasets for their easy and hard splits compared to the two baseline models.
DGRZSL outperformed the other models on easy splits when top-1 accuracy is used to evaluate the models.
The advantage of the model becomes more clear when the seen-unseen AUC metric is used as DGRZSL outperforms other models on most benchmarks.
The model is most successful on easy splits resulting in average improvements of 2.13\% and 2.73\% for top-1 accuracy and seen-unseen H, respectively.
CUB\_HARD is the only case where our method fails to improve upon the baslines.
Refer to \ref{fig:AUC} for the visualization of the seen-unseen curves for our model, GAZSL and CIZSL on all four benchmarks.

\section{Conclusion}
We introduced the Description Generator Regularized ZSL (DGRZSL) model.
DGRZSL includes an additional component which produces semantic representations of the underlying classes based on generated visual features.
Combined with an additional regularization, this encouraged the generated semantic representations to be consistent with the input to the visual feature generator for both seen and hallucinated classes. 
%
Our experiments showed that this modification improved the generalization performance over state-of-the-art generative ZSL models in terms of both top-1 accuracy and seen-unseen metrics.
Our evaluation on multiple benchmark datasets shows that the DGRZSL performs well for different types of semantic representation, including both textual-based and attribute-based class descriptions.

{\small
\bibliographystyle{ieee_fullname}
\bibliography{egbib}
}

\end{document}

%% file: figs/fig_model.tex
\begin{figure*}[t]
\centering
\includegraphics[scale=0.22]{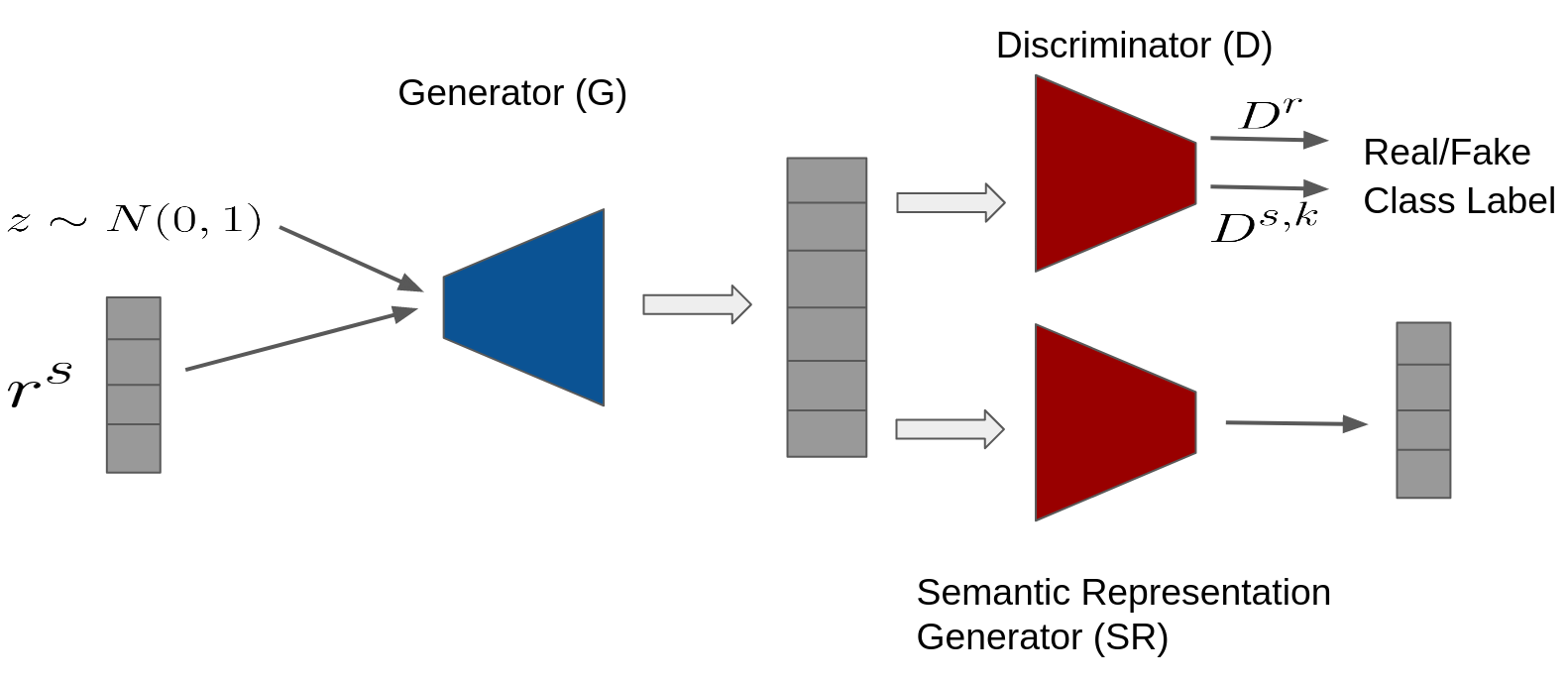}
\noindent\rule{17cm}{0.4pt}
\vspace{1cm}
\includegraphics[scale=0.22]{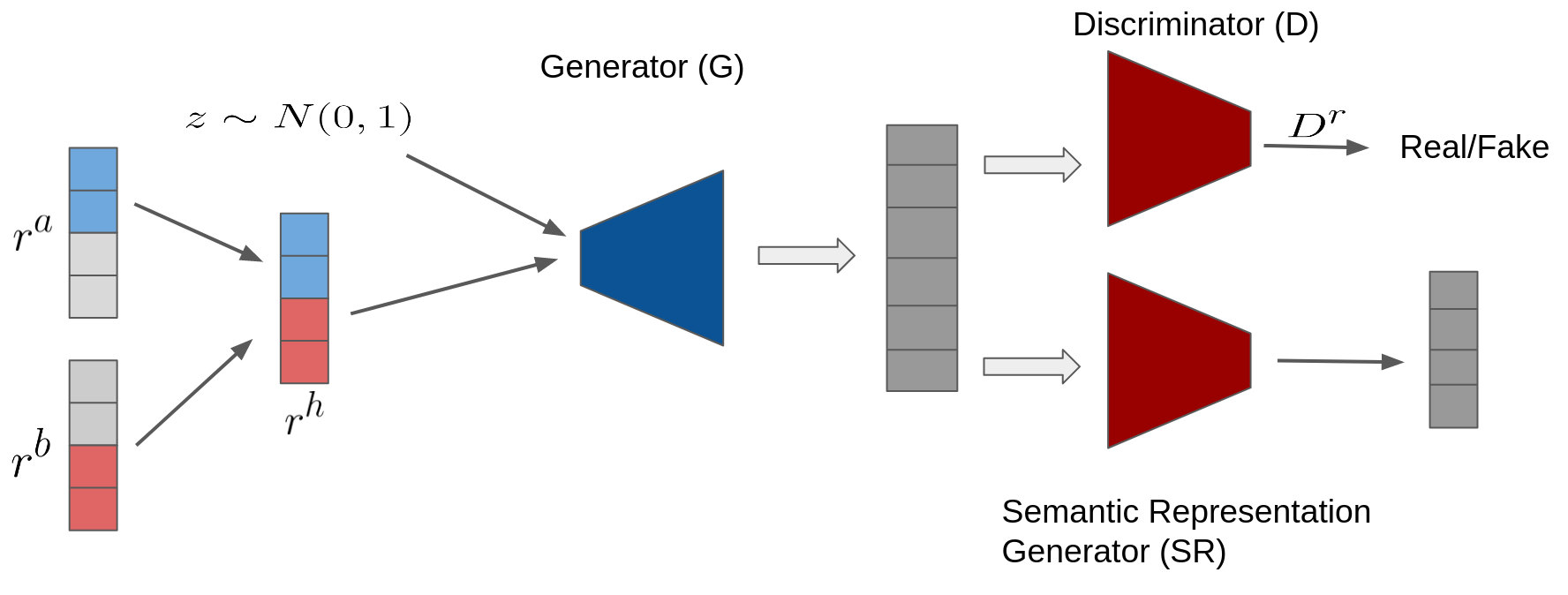}
\caption{Top image represents the model and its inputs when trained by following the conventional training process of GAN models. Bottom image represents the model and the input data when the model is trained by the hallucinated text introduced in \cite{Mohamed_Creative_2019}. The second head of the discriminator model which is responsible for classification is not used in this training process. Therefore, it is removed from the graph.}
\label{fig:model}
\end{figure*}

%% file: figs/tab_attribute_results_both_evaluation_methods.tex
\begin{table*}[t]
  \centering
  \begin{tabular}{c | c c c c c c c }
\cline{2-8}
  & Metric & \multicolumn{3}{c}{Top-1 Accuracy (\%)} & \multicolumn{3}{c}{Seen-Unseen H (\%)} \\
  Model Selection Set
  & Dataset & AWA2 & SUN & APY & AWA2 & SUN & APY\\
 \hline
 \multirow{3}{8em}{Validation (Ours)} & GAZSL \cite{Yizhe_ZSL_2018} & 56.33 & 60.76 & 27.18 & 28.36 & 25.59 & 14.77 \\
  & CIZSL \cite{Mohamed_Creative_2019} & 56.13 & 61.52& 30.98 & 26.66 & 26.27 & 16.31 \\
\cline{2-8}
 & DGRZSL & \textbf{57.79} & \textbf{61.94} & \textbf{38.59} & \textbf{33.50} & \textbf{26.70} & \textbf{18.09} \\
 \hline \hline
 \multirow{3}{8em}{Test (Original)} & GAZSL \cite{Yizhe_ZSL_2018} & 66.44 & 61.31 & \textbf{45.38} & 34.49 & 27.71 & 23.80 \\
  & CIZSL \cite{Mohamed_Creative_2019} & 68.81 & 61.25 & 39.50 & 34.72 & 27.12 & 23.85 \\
    \cline{2-8}
  & DGRZSL & \textbf{72.02} & \textbf{61.73}& 37.39& \textbf{47.53} & \textbf{28.02} & \textbf{28.4} \\
  \hline
\end{tabular}
  \caption{Zero-Shot classification results on AWA2, SUN and APY datasets. Our model outperforms both baseline models in all cases with an average top-1 accuracy increase of 9.4 \% and seen-unseen H increase of 17.26\% over CIZSL.  See text for more on evaluation method.}
  \label{tab:attribute_results_combined}
\end{table*}

%% file: figs/tab_textual_results_both_evaluation_methods.tex
\begin{table*}[t]
  \centering
  \begin{tabular}{ c |c c c c c c c c c }
\cline{2-10}
  & Metric & \multicolumn{4}{c}{Top-1 Accuracy (\%)} & \multicolumn{4}{c}{Seen-Unseen AUC (\%)} \\
  & Dataset & \multicolumn{2}{c}{CUB} &
 \multicolumn{2}{c}{NAB} & \multicolumn{2}{c}{CUB} & \multicolumn{2}{c}{NAB}\\
 Model Selection Set
 & Split-mode & 
 Easy & Hard & Easy & Hard & Easy & Hard & Easy & Hard \\
 \hline
 \multirow{3}{8em}{Validation (Ours)} & GAZSL \cite{Yizhe_ZSL_2018} & 42.40 & 18.83& 41.0 & \textbf{9.34} & 39.13 & \textbf{15.16} & 28.71 & 6.43 \\
  & CIZSL \cite{Mohamed_Creative_2019} & 41.38 & \textbf{18.99} & 40.56& 9.29 & 38.65 & 14.93 & 27.86 & 6.69 \\
 \cline{2-10}
 & DGRZSL & \textbf{42.44}& 17.36 & \textbf{42.71} & 8.76 & \textbf{39.77} & 11.70 & \textbf{29.81} & \textbf{6.87} \\
 \hline \hline
  \multirow{3}{8em}{Test (Original)} & GAZSL \cite{Yizhe_ZSL_2018} & 44.08& 14.46 & 36.36 & 8.74 & 39.69 & \textbf{12.49} & 24.68 & 6.48 \\
  & CIZSL \cite{Mohamed_Creative_2019} & 44.42& \textbf{14.8} & 36.85 & \textbf{9.04} & 39.46 & 11.85 & 24.83 & 6.48 \\
 \cline{2-10}
  & DGRZSL & \textbf{45.48} & 14.29 & \textbf{37.62} & 8.913 & \textbf{40.20} & 12.29 & \textbf{26.09} & \textbf{6.58} \\
 \hline
\end{tabular}
\caption{Zero-Shot classification results on CUB and NAB datasets. See text for more on evaluation method.}
  \label{tab:textual_results_combined}
\end{table*}

%% file: figs/fig_AUC.tex
\begin{figure}[t]
\centering
\includegraphics[width=0.49\textwidth]{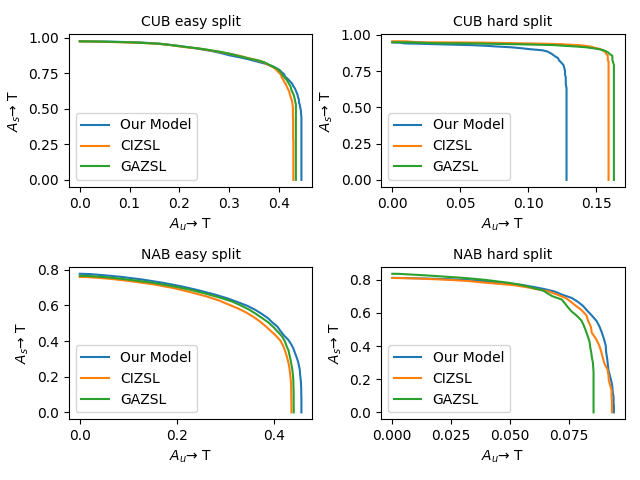}
\caption{Seen-Unseen accuracy curves with two splits for CUB and NAB datasets. Our model has a greater area under seen-unseen curve in three cases compare to the baselines demonstrating a better generalizability.}
\label{fig:AUC}
\end{figure}